%% file: _4pager.tex
\documentclass{article}
\usepackage{spconf,amsmath,epsfig}
\usepackage{booktabs}
\usepackage{multirow}
\usepackage{amsmath}
\usepackage{tabularx}
\usepackage{xcolor}
\usepackage{siunitx}
\usepackage{mathtools}
\usepackage{dsfont}
\usepackage[style=ieee, doi=false, url=false, isbn=false, maxnames=5, minnames=1, date=year, hyperref=true]{biblatex}
\usepackage{microtype}
\defbibheading{secbib}[\bibname]{%
  \centering{\bf\uppercase{#1}}
  \markboth{#1}{#1}
}
\renewbibmacro{in:}{}
\DeclareBibliographyDriver{unpublished}{%
    \usebibmacro{bibindex}%
    \usebibmacro{begentry}%
    \usebibmacro{author/translator+others}%
    \newunit
    \usebibmacro{title}%
    \newunit\newblock
    \usebibmacro{eprint}%
    \newunit\newblock
    \usebibmacro{date}
    \usebibmacro{finentry}%
}
\DeclareFieldFormat{eprint:arxiv}{%
  arXiv\addcolon\space
  \ifhyperref
    {\href{https://arxiv.org/abs/#1}{%
       \nolinkurl{#1}%
       \iffieldundef{eprintclass}
         {}
         }}
    {
    #1
     \iffieldundef{eprintclass}
       {}
       }
}

\usepackage[printonlyused]{acronym}
\usepackage{csquotes}
\usepackage{pifont}
\newcommand{\cmark}{\ding{51}}%
\newcommand{\xmark}{\ding{55}}%
\usepackage[hidelinks]{hyperref}
\usepackage[capitalise,noabbrev]{cleveref} 
\Crefname{figure}{Fig.}{Figs.}

\def\cpar{\hss\egroup\line\bgroup\hss}

\newcommand{\sn}[1]{\textcolor{red}{[SOURCE NEEDED]}}

\DeclareMathOperator{\attention}{Attention}
\DeclareMathOperator{\softmax}{Softmax}

\title{LiT-4-RSVQA: Lightweight Transformer-based Visual Question Answering in Remote Sensing}
\name{%
    Leonard Hackel$^{*\,1,3}$, Kai Norman Clasen$^{*\,1}$, Mahdyar Ravanbakhsh$^{2}$, Beg\"{u}m Demir$^{1,3}$
}

\address{
    $^1$Technische Universit{\"a}t Berlin, Germany
    $^2$Zalando SE, Berlin, Germany\\
    $^3$BIFOLD -- Berlin Institute for the Foundations of Learning and Data, Germany
}

\begin{document}
%
\maketitle
\def\thefootnote{*}\footnotetext{These authors contributed equally to this work}\def\thefootnote{\arabic{footnote}}

\begin{abstract}
\Ac{VQA} methods in \ac{RS} aim to answer natural language questions with respect to an \ac{RS} image.
Most of the existing methods require a large amount of computational resources, which limits their application in operational scenarios in \ac{RS}.
To address this issue, in this paper we present an effective \ac{lit4rsvqa} architecture for efficient and accurate \ac{VQA} in \ac{RS}.
Our architecture consists of: 
i) a lightweight text encoder module;
ii) a lightweight image encoder module; 
iii) a fusion module; 
and iv) a classification module. 
The experimental results obtained on a \ac{VQA} benchmark dataset demonstrate that our proposed \ac{lit4rsvqa} architecture provides accurate \ac{VQA} results while significantly reducing the computational requirements on the executing hardware. Our code is publicly available at \url{https://git.tu-berlin.de/rsim/lit4rsvqa}.
\end{abstract}

\acresetall


\input{acronyms}

\section{Introduction}
\label{sec:intro}
As a result of the increased volume of \ac{RS} image archives and the amount of information that can be extracted from them, the development of \ac{VQA} methods has recently become an important research topic in \ac{RS}.
In the task of \ac{VQA} in \ac{RS}, the user asks a question to a system in natural language concerning the content of \ac{RS} images. 
During the last years, several \ac{VQA} methods have been presented in \ac{RS}. 
As an example, \cite{Lobry2020} defines the \ac{VQA} task in \ac{RS} as a classification problem.
In this work, a ResNet-152 model is employed to extract features from \ac{RS} images, while skip-thought vectors are used to extract features from the text data. 
Then, the multi-modal features obtained through point-wise multiplication are classified using a \ac{MLP} as classification head.
In \cite{Lobry2020}, the first large-scale \ac{VQA} benchmark dataset for \ac{RS} based on low- and high-resolution \ac{RS} images and OpenStreetMap data is also introduced. 
In the \ac{VQA} method presented in \cite{chappuis2022language}, the \ac{RNN} based skip-thought vectors model for text feature extraction used in \cite{Lobry2020} is replaced with the attention-based \ac{BERT} \cite{Devlin2018}.
For the feature fusion, \cite{chappuis2022language} uses \ac{MUTAN}, which enables a richer and more meaningful interaction between features compared to a simple point-wise operation.
In \cite{Siebert2022}, a multi-modal transformer-based \ac{VBFusion} architecture is proposed.
\ac{VBFusion} uses a ResNet-152 architecture as a region proposal system and a Visual\ac{BERT} \cite{Li2019} model to learn the joint representation of image and text modalities instead of simply combining modality-specific representations. 
A similar approach is employed in \cite{Silva2022}, utilizing \ac{MMBERT} \cite{khare2021mmbert}.
This method also uses a ResNet-152 as a feature encoder. 
Instead of different regions of the input image, the backbone extracts five input representations at different resolutions and combines image and text features with a shallow \ac{MMBERT} encoder.
A comprehensive review of the current state-of-the-art of image-language models in \ac{RS} is presented in \cite{wen2023visionlanguage}.

All of the above-mentioned architectures provide good \ac{VQA} accuracies. 
However, they are associated with high computational complexity and \ac{PC}, and thus have limited capability to be applied for operational \ac{VQA} applications in \ac{RS}. 
As an example, \ac{VBFusion} has a \ac{PC} of 277 million and requires about 184 billion \acp{FLOP} per forward pass.
To overcome these limitations, in this paper, we investigate the effectiveness of lightweight transformer-based models for \ac{VQA} problems in \ac{RS} that require few parameters and are thus associated with low computational requirements.

\input{figures/fig_arch}

\section{Lightweight Transformer-based Models for VQA}
\label{sec:arch}
Given a triplet $(\mathbf{I},\mathbf{Q},\mathbf{A})$ of an image, question, and answer, a \ac{VQA} system provides the answer $\mathbf{A}$ given the image $\mathbf{I}$ and the question $\mathbf{Q}$ as the input.
In line with the \ac{RS} literature, we consider \ac{VQA} in \ac{RS} as a classification problem, where $\mathbf{A}$ is one of $n_A$ predefined cases.
In this paper, we study lightweight models for \ac{VQA} in \ac{RS}.
To this end, we introduce the \ac{lit4rsvqa} architecture that consists of four modules: 
i) a lightweight encoder module for the text modality;
ii) a lightweight encoder module for the image modality; 
iii) a fusion module; 
and iv) a classification module (see \cref{fig:arch}). 
In detail, the text encoder module is realized as one light\-weight transformer (\ac{BERT}$_\text{TINY}$), while the image encoder module is comprised of one of three light\-weight image transformers: 
i)~\ac{Deit} \cite{Touvron2020};
ii)~\ac{MobileViT} \cite{Mehta2021};
or iii)~\ac{XCiT} \cite{XCiT2021}.
We select these models based on their proven success from the literature and investigate their effectiveness in \ac{VQA} in \ac{RS}.
To the best of our knowledge, they are introduced for the first time in \ac{VQA} problems in \ac{RS}.
We combine BERT$_\text{TINY}$ with one of the above-mentioned image encoders, resulting in three \ac{VQA} model configurations. 

In general, text and image transformers are trained using \ac{MSA}, which is a mechanism that enables the model to focus on different parts (tokens) of an input sequence. 
Attention computes scores between each pair of input tokens through matrix multiplications and softmax operations, resulting in weights determining how much attention should be given to each token.
The attention weights are utilized to prioritize relevant parts of the input, enabling the model to focus on the most essential tokens.
The attention operation \cite{Vaswani2017} can be defined as:

\begin{align}
    \attention(Q,K,V) = \softmax\left(\frac{QK^\text{T}}{\sqrt{d}}\right) V,
\end{align}
where $Q$, $K$, and $V$ are query, key, and value, respectively. 
For self-attention, they are defined as $Q = XW_Q$, $K = XW_K$ and $V = XW_V$, where $W_Q$, $W_K$ and $W_V$ are linear projection matrices, and $X$ is a sequence of $d$-dimensional embeddings of the input.
For normalization, the attention weights $QK^\text{T}$ are scaled by the square root of the embedding dimension $d$.
To facilitate more diverse features, attention is computed in $A$ parallel attention heads.
$L$ transformer block layers are stacked to create a transformer network \cite{Vaswani2017, Dosovitskiy2020}.

To obtain representations for the text modality, we utilize \ac{BERT}$_\text{TINY}$ \cite{Turc2019}, which is a distilled version of \ac{BERT} \cite{Devlin2018}. 
\ac{BERT}$_\text{TINY}$ uses the same general architecture as \ac{BERT} but with fewer layers $L$, attention heads $A$, and a smaller embedding dimension $d$.
It uses \ac{MSA} and a \ac{FFN} to extract features from text tokens of the question input $\mathbf{Q}$.
\ac{BERT}$_\text{TINY}$ is pre-trained using a three-step method:
1) a large teacher model with more layers, attention heads, and larger embedding dimension is trained using \ac{MLM} and next sentence objectives;
2) \ac{BERT}$_\text{TINY}$ is pre-trained with \ac{MLM}; 
3) \ac{BERT}$_\text{TINY}$ is additionally pre-trained using knowledge distillation of the large teacher model from the first step. 
During the last step, the student model BERT$_\text{TINY}$ learns from the soft labels produced by the teacher model.

For the image feature extraction, \ac{Deit} \cite{Touvron2020} uses \ac{MSA} and a \ac{FFN}.
It extracts token embeddings from $\mathbf{I}$ by splitting the image into patches following \cite{Dosovitskiy2020}.
Similar to \ac{BERT}$_\text{TINY}$, it uses a reduced number of attention heads $A$ and a smaller embedding dimension $d$ to reduce the number of parameters and increase the throughput of the model.

\ac{MobileViT} \cite{Mehta2021} uses $n\times n$ and point-wise convolutions to encode local information about a pixel and project the information into a higher dimensional space.
The resulting tensors are reshaped (unfolded) into non-overlapping patches (tokens) similar to \cite{Dosovitskiy2020}, and attention is applied to the tokens to model long-range interactions.
Afterwards, the tokens are reshaped into their original tensor dimensions (folding), which is possible as the previous unfolding operation keeps the pixel and patch order intact.
This restores the original pixel location and allows for all pixels to encode information about all other pixels without the need for a large number of tokens in the attention operation.
Combined with dimensionality-reducing MobileNet-blocks, the unfolding-attention-folding operations reduce the latency and the number of parameters.

\ac{XCiT} \cite{XCiT2021} changes the order of matrix operations in \ac{MSA} and thus greatly reduces the cost of the attention operation.
This \ac{XCA} between keys and queries is calculated between channels instead of tokens, significantly reducing the number of
calculations required. 
\ac{XCA} \cite{XCiT2021} can be defined as:

\begin{align}
\attention_\text{XC} (Q, K, V) = V \softmax\left(\cfrac{\hat{K}^\text{T} \hat{Q}}{\tau}\right),
\end{align}
where $\hat{K}$ and $\hat{Q}$ are the normalized $K$ and $Q$, respectively, and $\tau$ is a learnable temperature parameter.
As this change removes explicit communication between patches, a block of convolutions, batch-norm, and non-linearity is added after each \ac{XCA} block to re-introduce information exchange.
For further detailed information on the considered models, we refer the reader to their respective papers mentioned above.

The feature fusion module consists of two linear projections and a modality combination.
The projections map the two modalities with dimensions $d_t$ and $d_v$ into a common dimension $d_f$, where $d_t$ and $d_v$ denote the dimensions of the flattened output of the text and image encoder modules, respectively.
The value of $d_v$ differs depending on the used lightweight transformer.
The projected features are then element-wise multiplied as in \cite{Lobry2020}. 
The classification module is defined as an \ac{MLP} projection head. 
After training of the proposed architecture, the \ac{VQA} system can be used to generate an answer $\mathbf{A}$ for a given input image $\mathbf{I}$ based on a natural language question $\mathbf{Q}$.

\section{Experimental Results}
The experiments were conducted on the RSVQAxBEN benchmark dataset \cite{Lobry2021}, which contains almost 15 million image/question/answer triplets extracted from the BigEarthNet-S2 dataset \cite{Sumbul2021}.
The questions provided in this dataset are about: i) the presence of one or more specific \ac{LULC} classes, where the answers are associated to \enquote{Yes/No} (called \emph{Yes/No}); 
and ii) the type of the \ac{LULC} classes, where the answers are one or more \ac{LULC} class names (called \emph{\ac{LULC}}).
In this paper, we used all available Sentinel-2 bands with \SI{10}{\meter} and \SI{20}{\meter} spatial resolution included in the BigEarthNet-S2 dataset, as suggested in \cite{Siebert2022}. We restricted the model output to the $n_A = 1,000$ most frequent answers.
We use the train/validation/test split as proposed in \cite{Lobry2021}.

In the experiments, we analyze our results between each other and compare them with those obtained from: 
i) VBFusion \cite{Siebert2022}; 
and ii) \ac{DBBT}, which is a \ac{VQA} model that exploits the large transformer Deit3$_\text{Base}$ \cite{Touvron2020} as image encoder and BERT$_\text{TINY}$ as a text encoder.
Each model (except VBFusion, for which we used the same implementation proposed in \cite{Siebert2022}) is evaluated under two training regimes: 
i) the image encoder is pre-trained for 100 epochs on BigEarthNet-S2, and fine-tuned for additional ten epochs on the RSVQAxBEN dataset \cite{Lobry2021}; 
and ii) the full \ac{VQA} network is trained in an end-to-end fashion for ten epochs. 
In both cases, the text encoders use pre-trained weights from Huggingface \cite{Wolf2020}.
Both training regimes use a linear-warmup-cosine-annuling learning rate schedule with a learning rate of \num{5e-4} after 10,000 warm-up steps with batch size and dropout set to 512 and 0.25, respectively. 
All models are trained on a single A100 GPU with matrix multiplication precision set to \enquote{medium} in Pytorch 1.13.1. 
To evaluate the results, the models are compared in terms of their: 
i) accuracy on the two question types Yes/No and \ac{LULC};
ii) \ac{OA}, which is the micro average of all answer classes; 
iii) \ac{AA}, which is the macro average of the two aforementioned question types; 
iv) \acf{PC};
and v) \acfp{FLOP}.

\Cref{tab:acc_comp} shows the experimental results.
From the table, one can see that all models within our \ac{lit4rsvqa} architecture provide competitive accuracies with significantly reduced computational complexity compared to both baseline models. 
Among the models in \ac{lit4rsvqa}, the configuration with pre-trained \ac{XCiT}  achieves the highest accuracy with less than one-tenth of the number of parameters and one-seventh of the computational effort measured in \acp{FLOP} compared to \ac{DBBT}.
With an \ac{AA} of \SI{64.61}{\percent}, it is almost \SI{9}{\percent} better than \ac{VBFusion} \cite{Siebert2022} and more than \SI{2.5}{\percent} better than \ac{DBBT}.
However, it is worth noting that \ac{XCiT} performs worst of all compared configurations when trained end-to-end.
In terms of \acp{FLOP}, the \ac{Deit}-based configuration demands the fewest computational resources.
It uses less than \SI{10}{\percent} of the \acp{FLOP} in comparison with the DBBT and more than 600 times fewer \acp{FLOP} when compared with \ac{VBFusion}. 
However, the accuracy of this model is higher in all evaluated metrics than the respective DBBT model and \ac{VBFusion}.
In addition, one can observe that:
i) models with pre-trained image encoders perform better than end-to-end trained networks;
and ii) better-performing pre-trained models do not correspond to better-performing end-to-end trained models, highlighting the advantage of using a pre-trained image encoder.

\input{figures/tab_results_without_sota_v2}



\section{Conclusion}
In this paper, we have studied efficient trans\-for\-mer-based models in the framework of \ac{VQA} in \ac{RS}.
In particular, we have introduced the LiT-4-RSVQA architecture.
Our architecture is based on lightweight transformer encoder modules, a feature fusion module, and a classification module.
Specifically, \ac{BERT}$_\text{TINY}$ is used as a lightweight text encoder, and \ac{Deit}, \ac{MobileViT}, and \ac{XCiT} are considered as lightweight image encoders.
Experimental results show that the investigated models can achieve high accuracy in the task of \ac{VQA} for \ac{RS} while having significantly fewer parameters and, therefore, lower computational requirements than larger models.
We would like to note that our architecture is not limited to the considered lightweight models. 
As future works, we plan to: i) investigate different encoder models and on-board processing with \acp{FPGA} and \acp{ASIC}; and ii) develop a library to simplify the development of \ac{VQA} models in \ac{RS}.

\section{Acknowledgements}
This work is supported by the European Research Council (ERC) through the ERC-2017-STG BigEarth Project under Grant 759764 and by the European Space Agency through the DA4DTE (Demonstrator precursor Digital Assistant interface for Digital Twin Earth) project and by the German Ministry for Economic Affairs and Climate Action through the AI-Cube Project under Grant 50EE2012B.



\section{REFERENCES}
\ninept
\printbibliography[heading=none]

\end{document}

%% file: acronyms.tex
\acrodef{DL}{deep learning}
\acrodef{ML}{machine Learning}
\acrodef{RS}{remote sensing}
\acrodef{ViT}{vision transformer}
\acrodef{ReLU}{rectified linear unit}
\acrodef{FCN}{fully convolutional network}
\acrodef{NLP}{natural language processing}
\acrodef{PC}{parameter count}
\acrodef{FLOP}{floating point operation}
\acrodef{GFLOP}[GF]{billion floating point operation}
\acrodef{VQA}{visual question answering}
\acrodef{EO}{Earth Observation}
\acrodef{lit4rsvqa}[LiT-4-RSVQA]{lightweight transformer-based \ac{VQA} in \ac{RS}}
\acrodef{RSVQA}{\ac{VQA} in \ac{RS}}
\acrodef{CNN}{convolutional neural network}
\acrodef{MS}{multispectral}
\acrodef{SAR}{synthetic aperture radar}
\acrodef{SA}{self-attention}
\acrodef{CV}{computer vision}
\acrodef{AA}{average accuracy}
\acrodef{OA}{overall accuracy}
\acrodef{MLP}{multi-layer perceptron}
\acrodef{MLM}{masked language modeling}
\acrodef{XCA}{cross-covariance attention}
\acrodef{LULC}{land use/land cover}
\acrodef{RNN}{recurrent neural network}
\acrodef{MSA}{multi-head self-attention}
\acrodef{FFN}{feed-forward network}
\acrodef{ASIC}{application-specific integrated circuit}
\acrodef{FPGA}{field-programmable gate array}
\acrodef{Deit}[Deit Tiny]{data-efficient vision transformer}
\acrodef{MobileViT}[MobileViT-S]{mobile vision transformer}
\acrodef{XCiT}[XCiT Nano]{cross-covariance image transformer}
\acrodef{MUTAN}{multi-modal tucker fusion for \ac{VQA}}
\acrodef{VBFusion}{Visual\ac{BERT} fusion}
\acrodef{BERT}{bidirectional encoder representations from transformers}
\acrodef{MMBERT}{multi-modal \ac{BERT}}
\acrodef{DBBT}{Deit3$_\text{Base}$ + \ac{BERT}$_\text{TINY}$}

%% file: figures/fig_arch.tex
\begin{figure*}
  \centering
  \includegraphics[width=0.7\textwidth]{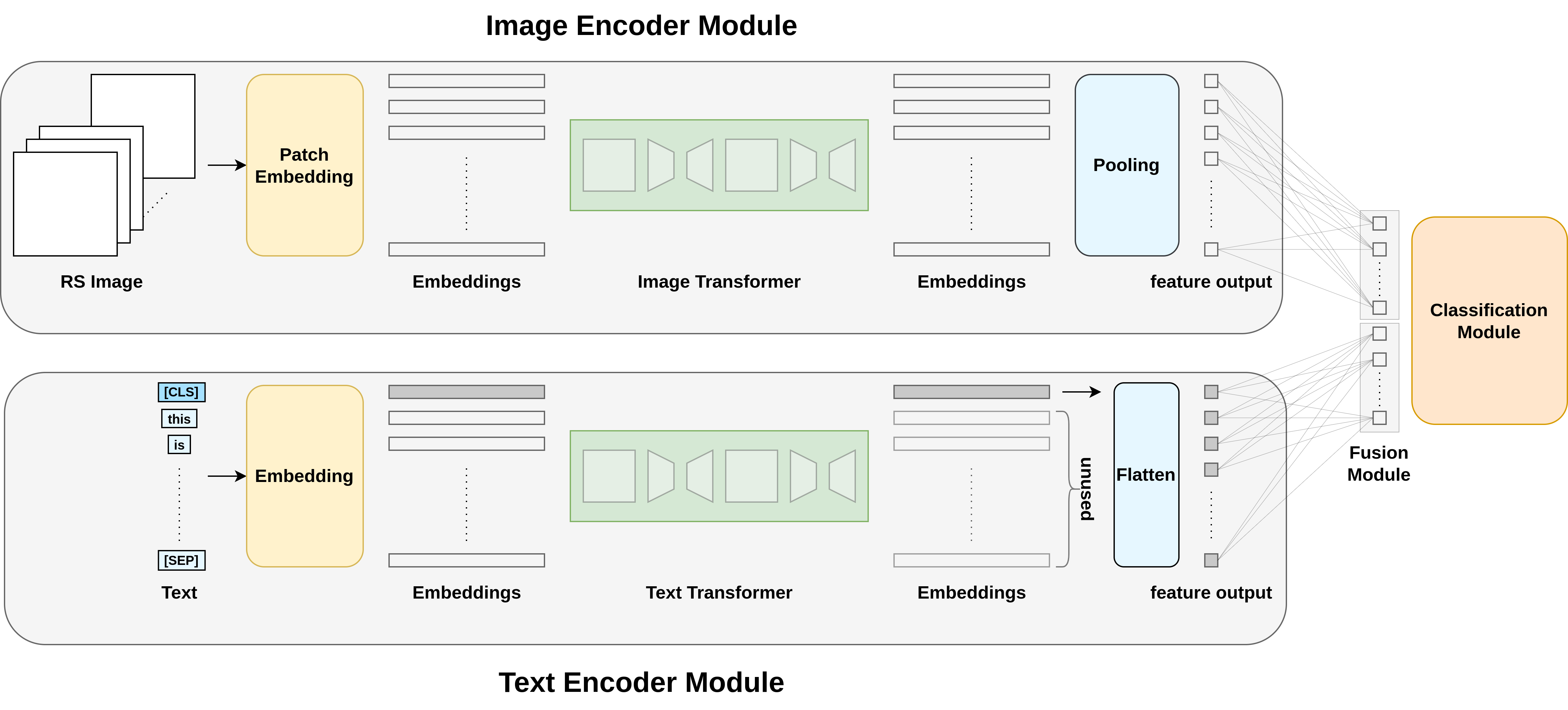}
  \caption{Illustration of the \acs{lit4rsvqa} architecture.}
  \label{fig:arch}
\end{figure*}

%% file: figures/tab_results_without_sota_v2.tex
\begin{table*}
    \centering
    \footnotesize
    \caption{Accuracy per question type as well as \acf*{AA} and \acf*{OA} obtained on the RSVQAxBEN dataset. \Acf*{PC} and  \acfp*{FLOP} are given in millions and billions, respectively. The image encoders use all available Sentinel-2 bands with 10m and 20m spatial resolution. Results are averaged over three runs, with standard deviations provided in brackets.
    \cmark$^{\mbox{\tiny{\ding{115}}}}$~Pre-trained~on~ImageNet; \cmark$^*$~pre-trained~on~BigEarthNet~\cite{Sumbul2021}. 
    }
    \label{tab:acc_comp}
    \begin{tabular}{llcccccccrr}
        \toprule
        \multirow{2}{*}{\textbf{Architecture}} & \multicolumn{3}{c}{\textbf{Encoder}} & \multirow{2}{*}{\textbf{Fusion}} & \multicolumn{2}{c}{\textbf{Question Type}} & \multirow{2}{*}{\textbf{\acs{AA}}} & \multirow{2}{*}{\textbf{\acs{OA}}} & \multirow{2}{*}{\textbf{\acs{PC}}} & \multirow{2}{*}{\textbf{\acsp{FLOP}}}  \\
        \cmidrule(lr){2-4}
        \cmidrule(lr){6-7}
        & Image & pre- & Text && LULC & Yes/No\\
        & & trained\\
        \toprule
        \multirow{1}{*}{VBFusion \cite{Siebert2022}}
        & ResNet-152 & \cmark$\mathrlap{^{\mbox{\tiny{\ding{115}}}}}$ & - & VisualBert & 26.26 & 86.56 & 55.80 & 76.10 & 277.0 & 183.9\\
        \cmidrule(lr){1-11}
        \multirow{2}{*}{DBBT} & Deit3 Base & \cmark$\mathrlap{^*}$ & Bert Tiny & Simple & 36.26 (1.10) & 87.83 (0.30) & 62.04 (0.70) & 79.15 (0.43) & 92.6 & 4.4\\
                                  & Deit3 Base & \xmark                & Bert Tiny & Simple & 29.53 (3.10) & 85.21 (1.39) & 57.36 (2.24) & 75.84 (1.67) & 92.6 & 4.4\\
        \cmidrule(lr){1-11}
        \multirow{6}{*}{LiT-4-RSVQA}
        & Deit Tiny    & \cmark$\mathrlap{^*}$ & Bert Tiny & Simple & 38.37 (1.17)          & 88.65 (0.39)          & 63.51 (0.64)          & 80.19 (0.39)          & 11.0         & \textbf{0.3}\\
        & MobileViT S & \cmark$\mathrlap{^*}$ & Bert Tiny & Simple & 34.43 (1.94)          & 89.02 (0.27)          & 61.37 (0.96)          & 79.95 (0.23)          & 10.4         & 0.5\\
        & XCiT Nano   & \cmark$\mathrlap{^*}$ & Bert Tiny & Simple & \textbf{39.50} (2.04) & \textbf{89.72} (0.65) & \textbf{64.61} (1.35) & \textbf{81.27} (0.88) & \textbf{8.1} & 0.6\\
        \cmidrule(lr){2-11}
        & Deit Tiny    & \xmark & Bert Tiny & Simple & 32.53 (1.93) & 86.80 (0.43) & 59.67 (1.16) & 77.67 (0.66) & 11.0         & \textbf{0.3}\\
        & MobileViT S & \xmark & Bert Tiny & Simple & 27.18 (3.18) & 85.51 (2.19) & 56.34 (2.63) & 75.70 (2.32) & 10.4         & 0.5\\
        & XCiT Nano   & \xmark & Bert Tiny & Simple & 21.45 (4.07) & 81.97 (4.66) & 51.71 (4.36) & 71.79 (4.56) & \textbf{8.1} & 0.6\\
        \bottomrule
    \end{tabular}
\end{table*}

%% file: references.bib
@article{Mehta2021,
    author = "Mehta, Sachin and Rastegari, Mohammad",
    journal = "ICLR",
    month = "10",
    title = "{MobileViT}: Light-weight, General-purpose, and Mobile-friendly Vision Transformer",
    url = "http://arxiv.org/abs/2110.02178",
    year = "2022"
}

@article{Dosovitskiy2020,
    author = "Dosovitskiy, Alexey and Beyer, Lucas and Kolesnikov, Alexander and Weissenborn, Dirk and Zhai, Xiaohua and Unterthiner, Thomas and Dehghani, Mostafa and Minderer, Matthias and Heigold, Georg and Gelly, Sylvain and Uszkoreit, Jakob and Houlsby, Neil",
    doi = "10.48550/arxiv.2010.11929",
    journal = "arXiv preprint 2010.11929",
    month = "10",
    title = "An Image is Worth 16x16 Words: Transformers for Image Recognition at Scale",
    url = "https://arxiv.org/abs/2010.11929v2",
    year = "2020"
}

@article{Vaswani2017,
    author = "Vaswani, Ashish and Shazeer, Noam and Parmar, Niki and Uszkoreit, Jakob and Jones, Llion and Gomez, Aidan N. and Kaiser, Lukasz and Polosukhin, Illia",
    journal = "NeurIPS",
    month = "6",
    title = "Attention Is All You Need",
    url = "http://arxiv.org/abs/1706.03762",
    volume = "30",
    year = "2017"
}

@inproceedings{Devlin2018,
    author = "Devlin, Jacob and Chang, Ming Wei and Lee, Kenton and Toutanova, Kristina",
    doi = "10.48550/arxiv.1810.04805",
    isbn = "9781950737130",
    journal = "Conference of the North American Chapter of the Association for Computational Linguistics: Human Language Technologies",
    booktitle = "Conference of the North American Chapter of the Association for Computational Linguistics: Human Language Technologies",
    month = "10",
    pages = "4171-4186",
    title = "{BERT}: Pre-training of Deep Bidirectional Transformers for Language Understanding",
    url = "https://arxiv.org/abs/1810.04805v2",
    year = "2018"
}

@article{wen2023visionlanguage,
    title={Vision-Language Models in Remote Sensing: Current Progress and Future Trends}, 
    author={Congcong Wen and Yuan Hu and Xiang Li and Zhenghang Yuan and Xiao Xiang Zhu},
    year={2023},
    journal = "arXiv preprint 2305.05726",
    primaryClass={cs.CV},
    url = "https://arxiv.org/abs/2305.05726",
    doi = "10.48550/arxiv.2305.05726"
}

@article{Lobry2020,
    author = "Lobry, Sylvain and Marcos, Diego and Murray, Jesse and Tuia, Devis",
    doi = "10.1109/tgrs.2020.2988782",
    issn = "15580644",
    issue = "12",
    journal = "IEEE TGRS",
    month = "3",
    pages = "8555-8566",
    title = "{RSVQA}: Visual Question Answering for Remote Sensing Data",
    url = "https://arxiv.org/abs/2003.07333v2",
    volume = "58",
    year = "2020"
}

@inproceedings{Lobry2021,
    author = "Lobry, Sylvain and Demir, Begum and Tuia, Devis",
    booktitle = "IEEE IGARSS",
    doi = "10.1109/IGARSS47720.2021.9553307",
    isbn = "978-1-6654-0369-6",
    journal = "IEEE IGARSS",
    month = "7",
    pages = "1218-1221",
    title = "{RSVQA} Meets {BigEarthNet}: A New, Large-Scale, Visual Question Answering Dataset for Remote Sensing",
    url = "https://ieeexplore.ieee.org/document/9553307/",
    year = "2021"
}

@inproceedings{khare2021mmbert,
  title={{MMBERT}: Multimodal Bert pre-training for improved medical {VQA}},
  author={Khare, Yash and Bagal, Viraj and Mathew, Minesh and Devi, Adithi and Priyakumar, U Deva and Jawahar, CV},
  booktitle={IEEE ISBI},
  pages={1033--1036},
  year={2021},
}

@article{Touvron2020,
    author = "Touvron, Hugo and Cord, Matthieu and Douze, Matthijs and Massa, Francisco and Sablayrolles, Alexandre and Jégou, Hervé and Ai, Facebook",
    doi = "10.48550/arxiv.2012.12877",
    journal = "arXiv preprint 2012.12877",
    month = "12",
    title = "Training data-efficient image transformers \& distillation through attention",
    url = "https://arxiv.org/abs/2012.12877v2",
    year = "2020"
}

@article{Siebert2022,
    author = "Siebert, Tim and Clasen, Kai Norman and Ravanbakhsh, Mahdyar and Demir, Begüm",
    journal = "SPIE Image and Signal Processing for Remote Sensing",
    month = "10",
    title = "Multi-Modal Fusion Transformer for Visual Question Answering in Remote Sensing",
    url = "http://arxiv.org/abs/2210.04510",
    year = "2022"
}

@inproceedings{Silva2022,
    author = "Silva, João Daniel and Magalhães, João and Tuia, Devis and Martins, Bruno",
    booktitle = "ACM SIGSPATIAL International Workshop on AI for Geographic Knowledge Discovery",
    city = "New York, NY, USA",
    doi = "10.1145/3557918.3565874",
    isbn = "9781450395328",
    journal = "ACM SIGSPATIAL International Workshop on AI for Geographic Knowledge Discovery",
    month = "11",
    pages = "40-49",
    title = "Remote sensing visual question answering with a self-attention multi-modal encoder",
    url = "https://dl.acm.org/doi/10.1145/3557918.3565874",
    year = "2022"
}

@article{XCiT2021,
    author = "El-Nouby, Alaaeldin and Touvron, Hugo and Caron, Mathilde and Bojanowski, Piotr and Douze, Matthijs and Joulin, Armand and Laptev, Ivan and Neverova, Natalia and Synnaeve, Gabriel and Verbeek, Jakob and Jegou, Hervé",
    doi = "10.48550/arxiv.2106.09681",
    isbn = "9781713845393",
    issn = "10495258",
    journal = "NeurIPS",
    month = "6",
    pages = "20014-20027",
    title = "{XCiT}: Cross-Covariance Image Transformers",
    url = "https://arxiv.org/abs/2106.09681v2",
    volume = "24",
    year = "2021"
}

@article{Turc2019,
    author = "Turc, Iulia and Chang, Ming-Wei and Lee, Kenton and Toutanova, Kristina",
    journal = "arXiv preprint 1908.08962",
    month = "8",
    title = "{Well-read} Students Learn Better: On the Importance of Pre-training Compact Models",
    url = "http://arxiv.org/abs/1908.08962",
    year = "2019"
}

@article{Sumbul2021,
    author = "Sumbul, Gencer and de Wall, Arne and Kreuziger, Tristan and Marcelino, Filipe and Costa, Hugo and Benevides, Pedro and Caetano, Mário and Demir, Begüm and Markl, Volker",
    doi = "10.1109/MGRS.2021.3089174",
    journal = "IEEE GRSM",
    month = "5",
    title = "{BigEarthNet-MM}: A Large Scale Multi-Modal Multi-Label Benchmark Archive for Remote Sensing Image Classification and Retrieval",
    url = "http://arxiv.org/abs/2105.07921 http://dx.doi.org/10.1109/MGRS.2021.3089174",
    year = "2021",
    volume={9},
    number={3},
    pages={174-180},
}

@article{Li2019,
    author = "Li, Liunian Harold and Yatskar, Mark and Yin, Da and Hsieh, Cho-Jui and Chang, Kai-Wei",
    journal = "arXiv preprint 1908.03557",
    month = "8",
    title = "{VisualBERT}: A Simple and Performant Baseline for Vision and Language",
    url = "http://arxiv.org/abs/1908.03557",
    year = "2019"
}

@inproceedings{Wolf2020,
    author = "Wolf, Thomas and Debut, Lysandre and Sanh, Victor and Chaumond, Julien and Delangue, Clement and Moi, Anthony and Cistac, Pierric and Rault, Tim and Louf, Remi and Funtowicz, Morgan and Davison, Joe and Shleifer, Sam and von Platen, Patrick and Ma, Clara and Jernite, Yacine and Plu, Julien and Xu, Canwen and Scao, Teven Le and Gugger, Sylvain and Drame, Mariama and Lhoest, Quentin and Rush, Alexander",
    booktitle = "Conference on Empirical Methods in Natural Language Processing: System Demonstrations",
    city = "Stroudsburg, PA, USA",
    doi = "10.18653/v1/2020.emnlp-demos.6",
    journal = "Conference on Empirical Methods in Natural Language Processing: System Demonstrations",
    month = "11",
    pages = "38-45",
    title = "Transformers: State-of-the-Art Natural Language Processing",
    url = "https://www.aclweb.org/anthology/2020.emnlp-demos.6",
    year = "2020"
}

@inproceedings{chappuis2022language,
  title={Language Transformers for Remote Sensing Visual Question Answering},
  author={Chappuis, Christel and Mendez, Vincent and Walt, Eliot and Lobry, Sylvain and Le Saux, Bertrand and Tuia, Devis},
  booktitle={IEEE IGARSS},
  pages={4855--4858},
  year={2022},
}
